\documentclass{llncs}
\usepackage{graphicx}
\usepackage{multicol}
\usepackage{amsmath}
\usepackage{makeidx}

\newtheorem{Def}{Definition}

\begin{document}
%
            % start of the contributions
%
\title{Fuzzy Object-Oriented Dynamic Networks. I}
\titlerunning{Fuzzy Object-Oriented Dynamic Networks. I}  % abbreviated title (for running head)
\author{D. A. Terletskyi\inst{1}, A. I. Provotar\inst{2}}
\authorrunning{} % abbreviated author list (for running head)
\tocauthor{}
\institute{Taras Shevchenko national University of Kyiv, Kyiv, 03680, Ukraine
\email{dmytro.terletskyi@gmail.com},\\
\and
University of Rzeszow, Rzeszow, 35--310, Poland\\
\email{aprowata1@bigmir.net},\\
}

\maketitle              % typeset the title of the contribution

\begin{abstract}
The concepts of fuzzy objects and their classes are described that make it possible to structurally represent knowledge about fuzzy and partially-defined objects and their classes. Operations over such objects and classes are also proposed that make it possible to obtain sets and new classes of fuzzy objects and also to model variations in object structures under the influence of external factors.

%The abstract should summarize the contents of the paper using at least 70 and at most 150 words. It will be set in 9-point
%font size and be inset 1.0 cm from the right and left margins. There will be two blank lines before and after the Abstract. \dots
\end{abstract}
\section*{Introduction}
At present, one of main problems facing scientists in the field of artificial intelligence (AI) is the development of intelligent information systems (IIS) to solve problems arising in different AI fields. The solution of such a problem is often reduced to heuristic programming that can yield good results, but, in this case, the complex solution of the problem is not provided. Systems based on knowledge representation models (KRMs) belong to largest classes of IISs. To date, the following models are well-known and actively used: semantic networks, conceptual graphs, frames, scripts, logical and production models, ontologies, etc. But, despite the use of some KRM or other as the base for an IIS, an implemented system consists of at least two levels, namely, the KRM level and level of its practical implementation. An implementation of a KRM often creates definite problems and complexities connected with interacting different levels of the system. In this connection, an object-oriented knowledge representation model was proposed in \cite{Terletskyi-1}, which can be integrated into object-oriented programming languages and thereby to unite the KRM itself and the language used for its implementation. This approach will allow one to get rid of some levels of abstraction and to partly simplify the structure of the system being developed and, hence, the development process itself.

The present article describes main ideas that form the basis for constructing fuzzy object-oriented dynamic networks, in particular, fuzzy objects and classes that allow one to structurally represent knowledge about objects that are unfuzzily specified, are fuzzy, or are incompletely defined and, at the same time, to classify them. Operations over fuzzy objects and classes of fuzzy objects are also considered with the help of which sets and new classes of fuzzy objects can be constructed and, thereby, new knowledge can be obtained.

\section*{Fuzzy Objects and Classes}
Objects can be considered to be both arbitrary things from our real life and results of using our imagination. In other words, an object is every thing that can be perceived in some way or other, i.e., can be sensed or imagined. It is obvious that each object (irrespective of its nature) has definite properties that are characteristic for it. For example, let us consider a natural number. It is obvious that it must be integer and larger than zero. It is these characteristic properties of natural numbers that allow one to distinguish them from other objects. In fact, in order to reveal whether the numbers $-1$, $4.67$, and $5$ are natural, it is necessary to check their properties, in particular, to check whether these numbers possess the same properties as natural numbers. As a result of verification, it becomes obvious that $-1$ and $4.67$ are not natural numbers. From this it is possible to draw the conclusion that a natural number is a crisply defined object. But, in addition to crisp objects, there also are other objects such as fuzzy and approximately or incompletely described objects. They arise in trying to recollect something, in describing our desires or fancies, in searching for something about which we know very little, etc. They mentally arise when we want to formalize some intuitive guess whose whole nature is fuzzy from the viewpoint of logic or mathematics. L.~A.~Zadeh was the first to propose a formalistic approach to the description of such objects \cite{Zadeh}. With time, this approach was transformed into a theory, and many results were obtained within the framework of this theory, in particular, in the field of constructing information systems that operate with fuzzy concepts. Taking into account the flexibility and efficiency of the approach proposed by L. Zadeh, we use here some of his ideas for formal definition of fuzzy objects.

In addition to the aforesaid concerning objects and their properties, we will pay attention to the following important point: properties of an object and the object itself are closely interconnected and cannot exist separately. Properties do not exist in themselves without an object since some object is their manifestation, and properties cannot be seen, understood, and even described without it. In turn, an object cannot exist without properties since their absence makes the construction or even a representation of the object impossible. It is obvious that this takes place for both crisp and fuzzy objects. Therefore, a formal definition of the concept of an object cannot be formulated if its properties are not formally defined and vice versa, it is impossible to define properties without knowing an object. In this connection, we first define properties of an object and then the object itself with allowance made for their interrelations.

Before considering properties of an object, note that they are subdivided into quantitative and qualitative ones.
\begin{Def}
A fuzzy quantitative property of an object $A$ is a tuple of the form $p(A)=(V(p(A)),u(p(A)))$, where $V(p(A))=\{v_1/\mu(v_1),\dots,v_n/\mu(v_n)\}$ is a fuzzy set describing the quantitative value of the property $p(A)$ and $u(p(A))$ are its measurement units.
\end{Def}
{\bf Example 1} Let us consider an object, for example, an apple; one of its properties is its mass. After weighing it, we obtain its exact mass, and this will determine the value of its crisp quantitative property described in \cite{Terletskyi-1}. But if it is impossible to approximately or exactly weigh an apple and thereby determine its mass, then we can represent it using a fuzzy set $p_m(A)=(V(p_m(A)),u(p_m(A)))$. If we take an apple and, based on our sensation, reveal that its mass is about $100\ g$, then this mass can be represented in the form
\[p_m(A)=(\{95/0.8+100/0.9+105/1+110/0.9+115/0.8\},\ g).\ \spadesuit\]

Let us define the equivalence of two fuzzy quantitative properties of objects to provide the possibility of comparing them.
\begin{Def}
Two fuzzy quantitative properties $p(A)$ and $p(B)$ are equivalent, i.e., $Eq(p(A),p(B))=1$, if and only if the following conditions are fulfilled:
\begin{enumerate}
\item $u(p(A))=u(p(B))$;
\item $m(v_i)-m(v_j)=0$, $i=\overline{1,n}$, $j=\overline{1,m}$;
\item $v_{k+1}-v_k=v_{w+1}-v_w$, $k=\overline{1,n-1}$, $w=\overline{1,m-1}$.
\end{enumerate}
\end{Def}
\begin{Def}
A fuzzy qualitative property of an object $A$ is a verification function $v_f(A):p(A)\rightarrow[0,1]$ that reflects the degree (measure) of truth (presence) of a property $p(A)$ for the object $A$.
\end{Def}

{\bf Example 2} Let us consider an object such as a water-melon and its property such as its (geometrical) form. The form of a water-melon can be described in different ways, but it is necessary to proceed from the fact that its form resembles a sphere. It is obvious that each concrete water-melon will have a unique form of a flat sphere. But an advantage of this approach is that a sphere is a crisply defined geometric object and, hence, the geometrical form of a water-melon, namely, the degree of its sphericity, can be represented as a function of verification of its sphericity, i.e., $vf_b(W):p_b(W)\rightarrow[0,1]$ $\spadesuit$.

Let us define the equivalence of two fuzzy qualitative properties of objects to provide the possibility of comparing them.
\begin{Def}
Two fuzzy qualitative properties $vf_i(A)$ and $vf_j(B)$ are equivalent, i.e., $Eq(vf_i(A),vf_j(B))=1$, if and only if the following condition is fulfilled: \[((vf_i(A)=vf_j(A))\cup(vf_i(B)=vf_j(B))).\]
\end{Def}
Since one object can have several properties that are quantitative and qualitative, it makes sense to define the concept of a fuzzy object specification.
\begin{Def}
A specification of a fuzzy object $A$ is a vector of the form $P(A)=(p_1(A),\dots,p_n(A))$, where $p_i(A)$, $i=\overline{1,n}$, is a quantitative or qualitative property of the object $A$.
\end{Def}
In other words, the specification of a fuzzy object can consist of crisply specified and fuzzy properties. Moreover, qualitative properties are special cases of fuzzy qualitative properties.

Next, using the concept of a fuzzy object specification, the concept itself of a fuzzy object can be directly defined.
\begin{Def}
A fuzzy object is a pair of the form $A/P(A)$, where $A$ is the object identifier and $P(A)$ is its specification.
\end{Def}

{\bf Example 3} We consider an object, namely, a water-melon that has, as is well known, a spherical form and assume that its weight is about $4\ kg$. Using the definition of a fuzzy object, it is possible to formally represent this water-melon as $W/P(W)$, where $P(W)=(vf_b(W),p_m(W))$, $vf_b(W)=0.8$, and \[p_m(W)=(\{3.7/0.8+4/0.9+4.3/1+4.5/0.9+4.7/0.7\},\ kg)\ \spadesuit.\]

It is obvious that the specification of the object W can have a larger number of properties, which depends on the level of detail to be taken into account in considering the object.

In addition to object properties, methods (operations) that can be applied to them should be taken into account, which makes it possible to operate with these objects to a certain degree. From this it makes sense to define the concept of a method of an object.
\begin{Def}
An operation (method) of a fuzzy object $A$ is a function $f(A)$ that can be applied to the object with allowance for distinctive features of its specification.
\end{Def}
Depending on the nature of an action on an object, methods can be divided into the following two types: modifiers, i.e., functions that can change the object, in particular, values of its properties, and exploiters, i.e., functions using objects as unchangeable parameters.

{\bf Example 4} Consider a fuzzy object, namely, a square $A$ that is introduced by its specification $P(A)=(p_1(A),\dots,p_4(A))$, where $p_1(A)=(4,sd.)$ is the number of its sides, $p_2(A)=(4,ang.)$ is the number of its angles, $p_3(A)=(\{2/0.9+2.2/1+2.4/0.9\},cm)$ is the size of its sides, and $p_4(A)=(90^\circ)$ is the degree measure of its angles. An example of an exploiter for the square $A$ is the function for computing its area, i.e., $S(A)=(v_i)^2$, where $v_i/\mu(v_i)\in V(p_3(A))$. As a result, we obtain
\[S(A)=\left(\left\{\frac{2^2}{0.9}+\frac{2.2^2}{1}+\frac{2.4^2}{0.9}\right\},cm\right)=\left(\left\{\frac{4}{0.9}+\frac{4.84}{1}+\frac{5.76}{0.9}\right\},cm\right).\]
In this case, an example of a modifier can be the following function of increasing a square whose application promotes the increase in the size of its sides: $H(A)=v_i+h$, where $h$ is some natural number. As a result, we obtain
\[H(A)=\left(\left\{\frac{2+h}{0.9}+\frac{2.2+h}{0.9}+\frac{2.4+h}{0.9}\right\},cm\right)\ \spadesuit.\]
Since several methods can be applied to the same object, it makes sense to define the concept of the signature of a fuzzy object.
\begin{Def}
A signature of a fuzzy object $A$ is a vector of the form $F(A)=(f_1(A),\dots,f_m(A))$, where $f_i(A)$, $i=\overline{1,m}$, is a method of the object $A$.
\end{Def}
Next, for purposes of comparison, we will define the equivalence of two fuzzy objects.
\begin{Def}
Two fuzzy objects $A$ and $B$ are considered to be of the same type if and only if they have equivalent specifications and the same methods can be applied to them, i.e., $P(A)=P(B)$ and $F(A)=F(B)$.
\end{Def}
It is obvious that each object, regardless of its nature, belongs to at least one class. In this connection, we will define the concept of a class of fuzzy objects.
\begin{Def}
A class of fuzzy objects is a tuple of the form $T=(P(T),F(T))$, where $P(T)$ is a specification when several fuzzy objects are described and $F(T)$ is their signature.
\end{Def}
By a class of fuzzy objects we understand properties of objects and methods that can be applied to them. In other words, a class of fuzzy objects is a generalized form of consideration of a number of fuzzy objects without the objects themselves. Analyzing the definition of a class of fuzzy objects, one can draw the following conclusion: in creating objects of this class, all fuzzy quantitative properties can be represented in the form of fuzzy sets of the second type \cite{Castillo}.

As well as in the case of classes of crisply specified objects, classes of fuzzy objects can be divided into two types, namely, homogeneous and heterogeneous. Reasons for this partition, distinctive features, and examples of these types of classes of objects are described in detail in \cite{Terletskyi-1}, \cite{Terletskyi-2}. Therefore, we pass directly to the definitions of the concepts of homogeneous and heterogeneous classes of fuzzy objects.
\begin{Def}
A homogeneous class of fuzzy objects is a fuzzy class of objects that describes only fuzzy objectsthe same type.
\end{Def}

{\bf Example 5} Any classes of convex polygons such as those of squares, rectangles, triangles, etc. are homogeneous classes of fuzzy objects $\spadesuit$.
\begin{Def}
A heterogeneous class of fuzzy objects is a tuple of the form $T=(Core(T),pr_1(A_1),\dots,pr_n(A_n))$, where $Core(T)=(P(T),F(T))$ is the core of the class of objects $T$ that consists of only the properties and methods that are common to specifications $P(A_1),\dots,P(A_n)$ and signatures $F(A_1),\dots,F(A_n)$, respectively; $pr_i(A_i)=(P(A_i),F(A_i))$, $i=\overline{1,n}$, are the projections of objects consisting of only the properties and methods that are peculiar to only fuzzy objects $A_1,\dots,A_n$.
\end{Def}

{\bf Example 6} Classes of all convex polygons, classes of all cars with the same brand name, the class of all TVs of one producer, etc. are heterogeneous classes of fuzzy objects $\spadesuit$.

\section*{Operations Over Fuzzy Objects and Classes}
A distinctive feature of all the methods mentioned above, irrespective of the class in which they are defined, is that all of them are local, i.e., are closed under the class in which they are defined. The reason is that methods of objects are intrinsically defined with allowance for specifications of objects, i.e., proceeding from properties. There are methods that are defined in one class of fuzzy objects but, at the same time, can be applied to other classes of fuzzy objects, i.e., they are polymorphous within certain limits. For implementing polymorphism, some object-oriented programming languages use the mechanism called operator overloading \cite{Stroustrup}.

In \cite{Terletskyi-2}, operations such as union, intersection, difference, symmetric difference, and cloning are considered that can be applied to any objects and classes of objects; at the same time, they do not require overloading and are rather universal. Important distinctive features of these methods are their set-theoretic (object-based) character and the possibility of obtaining new objects and classes of objects with their help. This is directly related to the important task RCG (Runtime Class Generation), i.e., the generation of classes during program execution. Despite the fact that all these operations were defined for crisply specified objects and classes of objects, they also can be applied to fuzzy objects and classes of fuzzy objects. Here, we will not define the operations themselves since they are similar for the case of fuzzy objects and classes of fuzzy objects. Let us consider an example of their practical application.

{\bf Example 7} Consider geometric figures, for example, a square $A$ and a rhombus $B$. It is obvious that they are representatives of different classes of convex polygons. We define their classes as follows:
\begin{gather*}
T(A)=(P(T(A)),F(T(A)))=((p_1(T(A)),\dots,p_6(T(A))),\\
(f_1(T(A)),f_2(T(A)))),\\
T(B)=(P(T(B)),F(T(B)))=((p_1(T(B)),\dots,p_6(T(B))),\\
(f_1(T(B)),f_2(T(B)))),
\end{gather*}
where $p_1(T(A))=(4,sd.)$ and $p_1(T(B))=(4,sd.)$ are the numbers of sides and $p_2(T(A))$ and $p_2(T(B))$ are the sizes of sides,
\begin{gather*}
p_2(T(A))=((\{2.9/0.95+3/1+3.4/0.75\},cm),\\
(\{2.9/0.95+3/1+3.4/0.75\},cm),(\{2.9/0.95+3/1+3.4/0.75\},cm),\\
(\{2.9/0.95+3/1+3.4/0.75\},cm)),\\
p_2(T(B))=((\{1.7/0.85+2/1+2.1/0.95\},cm),\\
(\{1.7/0.85+2/1+2.1/0.95\},cm),(\{1.7/0.85+2/1+2.1/0.95\},cm),\\
(\{1.7/0.85+2/1+2.1/0.95\},cm));
\end{gather*}
$p_3(T(A))=(4,ang.)$ and $p_3(T(B))=(4,ang.)$ are the numbers of angles of the figures; $p_4(T(B))$ and $p_4(T(A))=(90^\circ,90^\circ,90^\circ,90^\circ)$ are the grade measures of angles; $p_5(T(A))=1$ and $p_5(T(B))=1$ signify the equality of all sides; $p_6(T(A))=1$ and $p_6(T(B))=0.8$ signify the equality of all angles; $f_1(T(A))=4a$ and $f_1(T(B))=4b$ are methods for computing a perimeter; $f_2(T(A))=a^2$ and $f_2(T(B))=b^2\sin\alpha$ are methods for computing an area.

Analyzing the classes $T(A)$ and $T(B)$, one can state that they are classes of fuzzy squares and rhombuses, respectively, since their specifications contain fuzzy quantitative and qualitative properties.

Then we will define the following specifications and signatures of the fuzzy objects $A$ and B using the specifications and signatures of their classes:
\begin{gather*}
p_1(A)=4,\\
p_2(A)=((\{2.9/\{0.8/0.9+0.95/1+0.9/0.95\},3/\{0.9/0.9+1/1+0.85/0.85\},\\
3.4/\{0.7/0.95+0.75/1+0.6/0.8\}\},cm),(\{2.9/\{0.8/0.9+0.95/1+0.9/0.95\},\\
3/\{0.9/0.9+1/1+0.85/0.85\},3.4/\{0.7/0.95+0.75/1+0.6/0.8\}\},cm);\\
(\{2.9/\{0.8/0.9+0.95/1+0.9/0.95\},3/\{0.9/0.9+1/1+0.85/0.85\},\\
3.4/\{0.7/0.95+0.75/1+0.6/0.8\}\},cm),(\{2.9/\{0.8/0.9+0.95/1+0.9/0.95\},\\
3/\{0.9/0.9+1/1+0.85/0.85\},3.4/\{0.7/0.95+0.75/1+0.6/0.8\}\},cm);\\
p_3(A)=4;\ p_4(A)=(90^\circ,90^\circ,90^\circ,90^\circ);\ p_5(A)=1;\ p_6(A)=1;\\
p_1(B)=4;\\
p_2(B)=((\{1.7/\{0.7/0.8+0.85/1+0.9/0.95\},2/\{0.8/0.8+1/1+0.9/0.9\},\\
2.1/\{0.8/0.85+0.95/1+0.7/0.75\}\},cm),(\{1.7/\{0.7/0.8+0.85/1+0.9/0.95\},\\
2/\{0.8/0.8+1/1+0.9/0.9\},2.1/\{0.8/0.85+0.95/1+0.7/0.75\}\},cm),\\
(\{1.7/\{0.7/0.8+0.85/1+0.9/0.95\},2/\{0.8/0.8+1/1+0.9/0.9\},\\
2.1/\{0.8/0.85+0.95/1+0.7/0.75\}\},cm),(\{1.7/\{0.7/0.8+0.85/1+0.9/0.95\},\\
2/\{0.8/0.8+1/1+0.9/0.9\},2.1/\{0.8/0.85+0.95/1+0.7/0.75\}\},cm);\\
p_3(B)=4;\ p_4(B),p_5(B)=1;\ p_6(B)=0.8.
\end{gather*}

Using the fuzzy square $A$ and fuzzy rhombus $B$ as arguments, we will consider the operations of union, intersection, difference, symmetric difference, and cloning.

\emph{Union operation} $S=A/T(A)\cup B/T(B)=\{A,B\}/T(S)$.

As a result, a set of fuzzy objects $S$ and a new class of fuzzy objects $T(S)=(Core(T(S)),pr_1(A),pr_2(B))$ are obtained. Here,
\[Core(T(S))=(p_1(T(S)),p_2(T(S)),p_3(T(S)),f_1(T(S))),\]
where $p_1(T(S))=(4,sd.)$ is the number of sides; $p_2(T(S))=(4,ang.)$ is the number of angles, $p_3(T(S))=1$ signifies the equality of all sides, and $f_1(T(S))=4a$ signifies a method for computing perimeters;
\begin{gather*}
pr_1(A)=(p_2(A),p_4(A),p_6(A),f_2(A)),\\
pr_2(B)=(p_2(B),p_4(B),p_6(B),f_2(B)).
\end{gather*}
This allows us to draw the conclusion that $S$ is the set of fuzzy squares of type $T(A)$ and fuzzy rhombuses of type $T(B)$ and that the class $T(S)$ is a heterogeneous class of fuzzy objects and simultaneously describes two types of figures, namely, $T(A)$ and $T(B)$.

\emph{Intersection operation} $A/T(A)\cap B/T(B)=T(A\cap B)$.

As a result, a new class of fuzzy objects $T(A\cap B)=(Core(T(A\cap B)))$ is obtained, where
\[Core(T(A\cap B))=(p_1(T(A\cap B)),p_2(T(A\cap B)),p_3(T(A\cap B)),f_1(T(A\cap B)))\]
All the properties contained in $Core(T(A\cap B))$ coincide with the properties from the core of the class $Core(T(S))$ obtained as a result of union of the fuzzy objects $A$ and $B$. An analysis of the result does not allow one to exactly determine the type of geometric figures that describes the class of fuzzy objects $T(A\cap B)$ but, in this case, it may be said that it is homogeneous and consists of properties that are common to fuzzy squares of the class $T(A)$ and fuzzy rhombuses of the class $T(B)$.

\emph{Difference operation} $A/T(A)\setminus B/T(B)=T(A\setminus B)$.

As a result, we obtain a new class of fuzzy objects
\[T(A\setminus B)=(p_2(A),p_4(A),p_6(A),f_2(A)).\]
All properties of this class are exactly repeated in the projection of the object $A$ that is obtained as a result of union of the fuzzy objects $A$ and $B$, i.e.,
\[pr_1(A)=(p_2(A),p_4(A),p_6(A),f_2(A)).\]
It is obvious from the analysis of the result that, in contrast to the previous case, the homogeneous class of fuzzy objects $T(A\setminus B)$ describes a fuzzy rhombus and uses a smaller specification in this case.

\emph{Operation of symmetric difference} $A/T(A)\div B/T(B)=T(A\div B)$.

As a result, we obtain a new class of fuzzy objects $T(A\div B)=(pr_1(A),pr_2(B))$, where the projections
\begin{gather*}
pr_1(A)=(p_2(A),p_4(A),p_6(A),f_2(A)),\\
pr_2(B)=(p_2(B),p_4(B),p_6(B),f_2(B))
\end{gather*}
are similar to the projections obtained as a result of union of fuzzy objects $A$ and $B$. Analyzing the result, one can draw the conclusion that the heterogeneous class of fuzzy objects $T(A\div B)$ describes two types of geometric figures one of which can be a fuzzy rhombus and the second can be a fuzzy square or a rectangle.

\emph{Cloning operation} $Clone_1(A)=A_1/T(A)$.

As a result, a new numbered copy of the fuzzy object $A$ is obtained $\spadesuit$.

It is obvious from Example 7 that all the considered operations over fuzzy objects are exploiters since they do not change the fuzzy objects $A$ and $B$ but only use them as parameters. In this connection, we pass to the consideration of another type of operations over fuzzy objects, namely, modifiers. A detailed definition of a modifier of objects is given in \cite{Terletskyi-1}, \cite{Terletskyi-3} where five types of modifiers (complete, partial, generating, destroying, and replacing) are considered and the principle of construction of combined modifiers on their basis is shown. In \cite{Terletskyi-3}, the structures of objects and classes of objects are analyzed and also structural interrelations between properties of objects and classes of objects within the framework of their specifications are shown. These interrelations play an important role since they can be violated in the case of a modification and, as a result, the model does not necessarily correspond to real objects whose changes should be modeled. In \cite{Terletskyi-3}, this phenomenon is called the principle of reflection according to which changes in some property are rather frequently impossible in nature without a definite action on other properties that are connected with the former one. Example 7 partially demonstrates this principle since the properties $p_5(A)$ and $p_6(A)$ directly depend on values of the properties $p_2(A)$ and $p_4(A)$, respectively. It follows from this that a modification of the properties $p_2(A)$ and $p_4(A)$ sometimes stipulates a change in the properties $p_5(A)$ and $p_6(A)$ since, otherwise, this model will cease to describe the process being considered; moreover, the model becomes inconsistent. Assume that we modify the property $p_2(A)$ by specifying different lengths of figure sides. If we do not modify the property $p_5(A)$ in this case, then inconsistency will be obtained.

Processes of modification of crisply specified and fuzzy objects or classes of objects are underlain by the same principle, and a distinction lies only in the crispness or fuzziness of a specification. In this connection, we will give an example of modifications of a fuzzy object.

{\bf Example 8} We consider the fuzzy square $A$ from the previous example and modify it so that, as a result, a fuzzy rhombus is obtained. To this end, we will construct a partial modifier $M(A)=(m_2(p_2(A)),m_4(p_4(A)),m_6(p_6(A)))$. Here,
\begin{gather*}
m_2(p_2(A))=(\{2.3/\{0.8/0.9+0.95/1+0.9/0.95\},\\
2.6/\{0.9/0.9+1/1+0.85/0.85\},3.1/\{0.7/0.95+0.75/1+0.6/0.8\}\},cm),
\end{gather*}
$m_4(p_4(A))=(95^\circ,85^\circ,95^\circ,85^\circ)$, and $m_6(p_6(A))=0.85$ are functions of modification of the properties $p_2(A)$, $p_4(A)$, and $p_6(A)$, respectively. Thus, the fuzzy object $A$ is transformed into a fuzzy rhombus $A_1$ under the action of the modifier $M(A)$. It should be noted that this modification of the object $A$ also leads to a change in the signature of its class since the method $f_2(A)$ is incorrect for the fuzzy object $A_1$ and, therefore, the method $f_2(A_1)$ becomes undefined. Analyzing the result of this modification, one can draw the conclusion that a modification of objects and classes of objects stipulates the creation of new classes of objects, which also is directly related to RCG $\spadesuit$.

This example demonstrates only some of many aspects of the process of modification of objects and classes of objects. Note that all well-known operations over fuzzy sets can be used as functions for modifying fuzzy properties of objects. Let us consider the corresponding example.

{\bf Example 9} As an object, we use the fuzzy rhombus $B$ from Example 7. Consider the following partial modifiers:
\[M_1(B)=(m^1_6(p_6(B)))\ \text{and}\ M_2(B)=(m^2_6(p_6(B))),\]
where
\[m^1_6(p_6(B))=(v(p_6(B)))^{k^{=1}}= 0.8^{k^{-1}}\ \text{and}\ m^2_6(p_6(B))=(v(p_6(B)))^n=0.8^n\]
are functions for modifying the property $p_6(B)$ and $k$ and $n$ are natural numbers. In this case, the modifiers $M_1(B)$ and $M_2(B)$ are operations of dilution and concentration of fuzzy sets \cite{Rutkovskaya} $\spadesuit$.

\section*{Conclusions}
This article formulates definitions of a fuzzy object and a class of fuzzy objects that can underlie structural descriptions of fuzzy objects and, at the same time, allow for their classification. Two types of operations over fuzzy objects and classes of fuzzy objects (exploiters and modifiers) are also considered with the help of which sets and new classes of objects can be created and also changes in structures of objects and their properties can be modeled, in particular, under the influence of external factors. This approach allows one to model some capabilities of human intelligence, in particular, mechanisms of analysis, classification, search, and recognition of objects and classes of objects on the basis of their features. The results obtained allow us to testify that fuzzy object-oriented dynamic networks are constructed as models for the representation of knowledge on fuzzy objects, classes, and concepts.

%
% second contribution with nearly identical text,
% slightly changed contribution head (all entries
% appear as defaults), and modified bibliography
%

\end{document}